\documentclass[11pt,a4paper]{article}
\usepackage{acl2015}
\usepackage{times}
\usepackage{url}
\usepackage{latexsym}
\usepackage{amsmath}
\usepackage{graphicx}
\usepackage{color}

\title{Generate Image Descriptions based on Deep RNN and Memory Cells for Images Features}

\author{Shijian Tang \\
  Stanford University  \\
  {\tt sjtang@stanford.edu} \\\And
  Song Han \\
  Stanford University \\
  {\tt songhan@stanford.edu} \\}

\date{}

\begin{document}
\maketitle
\begin{abstract}


Generating natural language descriptions for images is a challenging task. The traditional way is to use the convolutional neural network (CNN) to extract image features, followed by recurrent neural network (RNN) to generate sentences. In this paper, we present a new model that added memory cells to gate the feeding of image features to the deep neural network. The intuition is enabling our model to memorize how much information from images should be fed at each stage of the RNN. Experiments on Flickr8K and Flickr30K datasets showed that our model outperforms other state-of-the-art models with higher BLEU scores.


\end{abstract}

\section{Introduction}

Generating natural language descriptions for images has became an attractive research topic in recent years. The task is to generate sentences or phrases to summarize and describe the contents shown in images. With this technique, the machines are enabled to imitate the behaviour of human beings who are able to capture the semantic meaning encoded in images. Some previous work from \newcite{Gupta:12}, \newcite{Kulkarni:11} and \newcite{Elliott:13} designed templates for the sentence descriptions. The task is to fill in the templates based on the images. However, these approaches strongly limited the capability of models to generate sentence descriptions to only fixed patterns. Other approaches transfer this task into a multimodal embedding problem. These work from \newcite{Farhadi:10}, \newcite{Jia:11}, \newcite{Socher:14}, \newcite{Ordonez:11} overlap with the scope of information retrieval. The goal is to map the images with sentences appearing in the training dataset together in a multimodal space. However, these models are only capable of returning sentence descriptions that existed in the training dataset.

Most of the state-of-the-art approaches are based on neural networks. These work combined convolutional neural network (CNN) with recurrent neural network (RNN) to generate image descriptions. \newcite{Karpathy:15} develop a multimodal RNN for this task. In this neural network, the image features extracted from the VGGNet (a pretrained CNN proposed in \newcite{Simonyan:14}) are fed into a RNN. Conditioned on the image features and previous words, the RNN will generate a sequence of words recurrently to describe the images. Similar to Kaparthy's work, Vinyals \newcite{Vinyals:14} used the GoogLeNet CNN to extract image features and train a LSTM (in \newcite{Hochreiter:97}) as sequence generator. \newcite{Mao:14} report a deep complex multimodal RNN for sentence generation.

In our approach, VGGNet is employed to extract image features and a deep multilayer RNN is chosen as a sequence generator, on top of which we informatively added memory gate that controls image feeding. In each time step of RNN \footnote{In the RNN language model, the time step is defined as the position of word in sentence.}, we feed word in current time step as well as the image features into the hidden layer of RNN. Inspired by the ideas in \newcite{Rolls:02} that the visual perception depends on short-term memory and has the recurrent natural, a memory gate is designed to control the input of image features to the hidden layer. The output of memory gate depends on the output of hidden layer at the previous time step. Before feeding into the hidden layer, the image features are multiplied by the output of gate element-wisely. Therefore, the memory gates act as memory cells for image features. Our model is trained on the Flickr8K and Flickr30K datasets from \newcite{Hodosh:13}. We evaluate the BLEU score (proposed in \newcite{Papineni:02}) of our model on the test datasets of both Flickr8K and Flickr30K. The preliminary results show that the performance of our model outperforms the state-of-the-art work.

\section{The Architecture of Model}

\subsection{Image Features Representation}

CNN has been proved as a powerful tool to extract image features, and has been widely used in image classification (\newcite{Krizhevsky:12}), object detection (\newcite{Girshick:14}) and other tasks. In this paper, We select the deep and powerful VGGNet to extract image features. Specifically, each raw image is fed into the VGGNet as input. After the forward propagation, the last fully-connected layer will output a 4096 dimensions vector as the image features for each image.

\subsection{Sentence Representation}

The sentence can be represented as a sequence of single word. The time step $t$ is defined as the index of $t$ th word in the sentence to represent the position of each word. Suppose the sentence contains $T$ words, the time step of first word is $t=1$, the second word is $t=2$, and for the last word is $t=T$. For each sentence, we add a special START token at the first time step to indicate the start of the sentence, as well as the END token at the last time step as the end of each sentence.

The single word is represented as a vector. Some pretrained word vector models have been developed such as word2vec by \newcite{Mikolov:13} and Glove by \newcite{Pennington:14}. However, in this model, we trained the word vectors from scratch instead of directly adopted the pretrained model, since generally the retrained word vector will achieve higher performance for specific task.

The standard RNN model can be expressed as,
\begin{align}\label{eq: std_rnn}
h(t) &= f(W^{s}x(t)+W^{h}h(t-1)+b^h)\\
y(t) &= softmax(W^dh(t)+b^d)\nonumber\\
\nonumber
\end{align}
where $h(t)$ is the output of the hidden layer at time step $t$, $x(t)$ is the word vector for the word at $t$, $h(t-1)$ is the output of hidden layer for the previous time step $t-1$, $f()$ is the activation function. $W^s$ has the dimension of $H$ by $D$ where $H$ is the dimension of hidden layer and $D$ is the dimension of word vector. $W^h$ has the dimension of $H$ by $H$. $W^d$ has the dimension of $V$ by $H$ with $V$ as the vocabulary size. $b^d$ and $b_h$ are the bias terms. Vector $y(t)$ represents the probability of each word in the vocabulary to be the next word conditioned on the input words from time step $1$ to $t$.

In our model, to improve the model capacity, we increase the depth of RNN by adding multiple hidden layers which is the same as deep transition RNN (DT-RNN) model reported by \newcite{Pascanu:14}. \newcite{Pascanu:14} shows that the DT-RNN is able to increase the size of family of functions it can represent in language modeling. Unlike the standard RNN in equation~\ref{eq: std_rnn} with only a single hidden layer at each time step, $N$ hidden layers are stacked together at each time step in DT-RNN. The forward propagation of this deep RNN model is,
\begin{align}\label{eq: rnn}
h_1(t) &= f(W^{s}x(t)+W^{h}h_N(t-1)+b^h) \nonumber\\
h_2(t) &= f(W^hh_1(t)+b^h)\nonumber\\
&...\\
h_N(t) &=f(W^hh_{N-1}(t)+b^h)\nonumber\\
y(t) &=softmax(W^dh_N(t)+b^d)\nonumber\\
\nonumber
\end{align}
where $h_1(t), h_2(t),..., h_N(t)$ represent the output of $N$ hidden layers at $t$. The word vector $x(t)$ as well as the output of last hidden layer at previous time step $h_{N}(t-1)$ are fed into the first hidden layer $h_1(t)$ at $t$. Then, output of current hidden layer feeds into the next hidden layer consecutively. The output $y(t)$ depends on the output of last hidden layer at current time step $h_N(t)$. In our model, the deep RNN in equation~\ref{eq: rnn} is chosen as the sequence learner of sentences.

\subsection{Memory Cells for Image Features}

We consider how to control feeding the image features into the deep RNN. Instead of feeding the image features directly, we add a gate to control the magnitude of image feature feeds. The value of the gate depends on the state of hidden layers at previous time step.
\begin{align}\label{eq: memory}
h_1(t) &=f(W^sx(t)+W^hh_N(t-1)+\nonumber\\
       & g(t)\circ (W^iCNN(I)+b^i)+b^h)\\
g(t) &=\sigma(W^gh_N(t-1)+b^g)\nonumber\\
\nonumber
\end{align}
where $I$ represents the raw image, and $CNN(I)$ is the image features extracted by CNN. $W^i$ has the dimension of $H$ by 4096 which maps the image features to the same space of hidden layers of RNN. $g(t)$ is the output of gate, and $\circ$ is the element-wise multiplication. $W^g$ transfers the value of the last hidden layer in the previous time step ($h_N(t-1)$ in the equation) to the gate $g(t)$. $b_i$ and $b_g$ are the bias terms. Here we use the $\sigma$ activation function and the value of $g(t)$ ranges from $0$ to $1$.

Based on equation~\ref{eq: memory}, the image features are fed into the first hidden layer at each time step, multiplied by the output of gate. Since the value of gate depends on the last hidden layer of previous time step, the gate controls how much information from image is still needed for the current time step. In the case of $g(t)=0$, the image features are not fed into RNN, while for $g(t)=1$, we feed full image features at each time step.

\begin{figure}[h]
\centering
\includegraphics[width=3in]{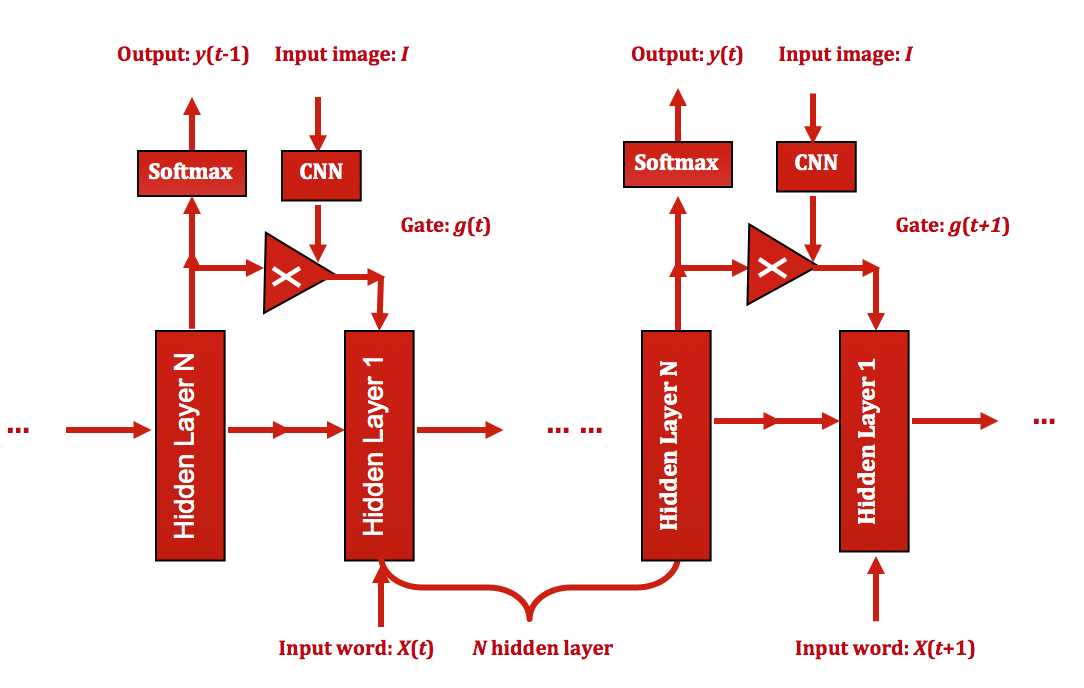}
\caption{The architecture of this model.}
\label{fig: model}
\end{figure}

Combining equations \ref{eq: rnn} and \ref{eq: memory} together, this model can be represented as:
\begin{align}\label{eq: model}
g(t) &=\sigma(W^gh_N(t-1)+b^g)\nonumber\\
h_1(t) &=f(W^sx(t)+W^hh_N(t-1)+\nonumber\\
&g(t)\circ (W^iCNN(I)+b^i)+b^h)\nonumber\\
h_2(t) &= f(W^hh_1(t)+b^h)\nonumber\\
&...\\
h_N(t) &=f(W^hh_{N-1}(t)+b^h)\nonumber\\
y(t) &=softmax(W^dh_N(t)+b^d)\nonumber\\
\nonumber
\end{align}
Figure~\ref{fig: model} shows the architecture of this model.

Recall the work of \newcite{Karpathy:15}, image features are only fed at the first time step of RNN. Due to the vanishing gradient problem, image features will not be learned well with long sentence and deep network. However, our model feed image features into RNN at each time step. Therefore, our model is still able to learn information from the image even for larger time steps. The magnitude of image features is conditioned on the hidden state of previous time step. In another word, the image features are encoded based on the status of how well our model has learned.

Compared with other work of \newcite{Vinyals:14} based on LSTM and work of \newcite{Mao:14} based on multimodal embeddings, our model has the advantage of lower model complexity and easier to train.

\section{Experiments}

\subsection{Dataset}

We experimented on the Flickr8K and Flickr30K datasets introduced in \newcite{Hodosh:13}. Each image in these datasets is described by $5$ independent sentences. Therefore, for each image, we can create $5$ samples with each one as an image-sentence pair. We have $8000$ and $31000$ images for Flickr8K and Flickr30K respectively. Each dataset has been splited into development data with $1000$ images, test data with $1000$ images and the rest images as training data. The data preprocessing procedure is the same as the work of \newcite{Karpathy:15}.

\subsection{Training}

During training, cross entropy loss was chosen as the loss function. Stochastic gradient descent (SGD) with minibatch size of $100$ image-sentence pairs was used during training. To make the model converge faster, RMSprop annealing policy \newcite{Hinton:12} was adopted, where the step size of each parameter is scaled by the window-averaged norm of its gradient.

To overcome the vanishing gradient problem, ReLU is chosen as the activation function. Also, we adopted the element-wise clip gradient tricks, where we clipped the gradient to $5$. To regularize the model, we add L2 norm of weights to the loss function, and as \newcite{Zaremba:14} suggested, we used dropout ratio of $0.5$ to all the layers except for the hidden layers.

As equation~\ref{eq: model} indicates, a model with large $N$ has deeper hidden layers, which leads to a large capacity. Considering the size of the dataset is not large and in order to prevent overfitting, we adopt a small $N=2$ with 2 hidden layers in the experiments in equation~\ref{eq: model}.

We find $50$ epochs are enough to train this model for both datasets, and the hidden size was tuned to $512$ to achieve the best performance.

\subsection{Generate Image Description}

The sentence description for each image in test dataset is generated by feeding the image features into the trained model with a START token. At each time step, we can directly choose the word corresponds to the one with highest probability in vocabulary as the output word, which is also the input word of next time step. Following this method, we can generate a sentence recurrently until we reach the END token.

To evaluate the performance, we use the BLEU score as evaluation metrics which has been widely adopted in the papers focus on this topic (\newcite{Karpathy:15}, \newcite{Vinyals:14}, \newcite{Mao:14}). The BLEU score will evaluate the similarity of the generated sentences with the ground truth sentences. Table~\ref{tab:flickr8k} and Table~\ref{tab:flickr30k} show the BLEU score for several models.

\begin{table}[h]
\begin{center}
\begin{tabular}{|l|rlll|}
\hline \bf Model & \bf B-1 & \bf B-2 & \bf B-3 & \bf B-4\\ \hline
\small{Our Model} & \bf 58.3 & \bf 39.7 & \bf 25.7 & \bf 16.6\\
\small{\newcite{Karpathy:15}} & 57.9 & 38.3 & 24.5 & 16.0\\
\small{\newcite{Mao:14}} & 57.8 & 27.5 & 23.1 & ---\\
\small{\newcite{Vinyals:14}} & 63.0 & 41.0 & 27.0 & --- \\
\small{Vinyals' net on VGGNet} & 58.2 & 37.8 & 19.0 & ---\\
\hline
\end{tabular}
\end{center}
\caption{\label{tab:flickr8k} The BLEU score on Flickr8K for different models. {\bf{B-n}} is the BLEU score up to n-gram.}
\end{table}

\begin{table}[h]
\begin{center}
\begin{tabular}{|l|rlll|}
\hline \bf Model & \bf B-1 & \bf B-2 & \bf B-3 & \bf B-4\\ \hline
\small{Our Model} & \bf 59.0 & \bf 39.5 & \bf 26.0 & \bf 17.1\\
\small{\newcite{Karpathy:15}} & 57.3 & 36.9 & 24.0 & 15.7\\
\small{\newcite{Mao:14}} & 54.8 & 23.9 & 19.5 & ---\\
\small{\newcite{Vinyals:14}} & 66.3 & 42.3 & 27.7 & 18.3 \\
\small{Vinyals' net on VGGNet} & --- & --- & --- & ---\\
\hline
\end{tabular}
\end{center}
\caption{\label{tab:flickr30k} The BLEU score on Flickr30K for different models. {\bf{B-n}} is the BLEU score up to n-gram.}
\end{table}

As shown on Table~\ref{tab:flickr8k} and Table~\ref{tab:flickr30k}, our model outperforms the results from \newcite{Karpathy:15} and \newcite{Mao:14}. While the performance of our model is lower than the original work from \newcite{Vinyals:14}. However, this is because in the original work of \newcite{Vinyals:14}, the authors used the GoogleNet (in \newcite{Szegedy:14}) to extract the image features, while we used VGGNet. Therefore, it is unfair to directly compare the BLEU score of our model with results reported by \newcite{Vinyals:14}.

To make a fair comparison with the network in \newcite{Vinyals:14}, we have downloaded the reproduced version of Vinyals' model from \url{http://cs.stanford.edu/people/karpathy/neuraltalk/}. In this reproduced model trained on Flickr8K, the image features feed into Vinyals' model are extracted by VGGNet, which is the same as the case in our model. From the last row of Table~\ref{tab:flickr8k}, we can find that the performance of our model is better than the model in \newcite{Vinyals:14} if both models use the VGGNet image features. Note that even though the reproduced model of \newcite{Vinyals:14} based on Flickr30K dataset is unavailable now, our model still outperforms other state-of-the-art works.

We also tried to feed image features only at first time step (i.e., set $g(t)=0$ except for the first time step) as well as feed full image features at each time step (i,e., set $g(t)=1$ for all time steps). But the results show that the performance all of these two schemes are lower than feeding image features at each time step with memory cells.

\section{Conclusion}
In this paper, we developed a new model for generating image descriptions. The image features extracted from VGGNet are fed into each time step of a multilayer deep RNN, where the image features vector is element-wisely multiplied by a memory vector determined by the state of the hidden layer at previous time step. Experiments on Flickr8K and Flickr30K datasets show that this model achieves higher performance on BLEU score. Our model also benefit from its low complexity and ease of training.

As the extension of this work, we will train our model on a larger dataset such as MSCOCO, and will increase the number of hidden layers at each time step to further improve the performance of our model. We will also try to adopt other CNNs such as GoogleNet to extract image features. Also, in this work, we do not fine-tune the CNNs on the new datasets, in future, we will try to train the model and tune the CNNs together.


\begin{thebibliography}{}

\bibitem[\protect\citename{Gupta and Mannem}2012]{Gupta:12}
Ankush Gupta and Prashanth Mannem.
\newblock 2012.
\newblock From image annotation to image
description.
\newblock In {\it NIPS}.

\bibitem[\protect\citename{Kulkarni \bgroup et al. \egroup}2011]{Kulkarni:11}
Girish Kulkarni, Visruth Premraj, Sagnik Dhar, Siming Li,
Yejin Choi, Alexander Berg, and Tamara Berg.
\newblock 2011.
\newblock Baby talk: Understanding and generating
simple image descriptions.
\newblock In {\it CVPR}.

\bibitem[\protect\citename{Desmond Elliott and Frank Keller}2013]{Elliott:13}
Desmond Elliott and Frank Keller.
\newblock 2013.
\newblock Image description using visual dependency representations.
\newblock In {\it EMNLP}.

\bibitem[\protect\citename{Farhadi \bgroup et al. \egroup}2011]{Farhadi:10}
Ali Farhadi, Mohsen Hejrati, Mohammad Amin Sadeghi, Peter Young, Cyrus Rashtchian, Julia Hockenmaier, and David Forsyth.
\newblock 2010.
\newblock Every picture tells a story: Generating sentences from images.
\newblock In {\it ECCV}.

\bibitem[\protect\citename{Jia \bgroup et al. \egroup}2011]{Jia:11}
Yangqing Jia, Mathieu Salzmann, and Trevor Darrell.
\newblock 2011.
\newblock Learning cross-modality similarity for multinomial data.
\newblock In {\it ICCV}.

\bibitem[\protect\citename{Socher \bgroup et al. \egroup}2011]{Socher:14}
Richard Socher, Andrej Karpathy, Quoc V. Le, Christopher D. Manning, and Andrew Y. Ng.
\newblock 2014.
\newblock Grounded compositional semantics for finding and describing images with sentences.
\newblock In {\it TACL}.

\bibitem[\protect\citename{Ordonez \bgroup et al. \egroup}2011]{Ordonez:11}
Vicente Ordonez, Girish Kulkarni, and Tamara L. Berg.
\newblock 2011.
\newblock Im2text: Describing images using 1 million captioned photographs.
\newblock In {\it NIPS}.

\bibitem[\protect\citename{Karpathy}2015]{Karpathy:15}
Andrej Karpathy and Li Fei-Fei.
\newblock 2015.
\newblock Deep visual-semantic alignments for generating image descriptions.
\newblock In {\it CVPR}.

\bibitem[\protect\citename{Simonyan and Zisserman}2014]{Simonyan:14}
Karen Simonyan and Andrew Zisserman.
\newblock 2014.
\newblock Very deep convolutional networks for large-scale image recognition.
\newblock {\it arXiv preprint arXiv:1409.1556}.

\bibitem[\protect\citename{Vinyals \bgroup et al. \egroup}2014]{Vinyals:14}
Oriol Vinyals, Alexander Toshev, Samy Bengio, and Dumitru Erhan.
\newblock 2014.
\newblock Show and tell: A neural image caption generator.
\newblock {\it arXiv preprint arXiv:1411.4555}.

\bibitem[\protect\citename{Hochreiter and Schmidhuber}1997]{Hochreiter:97}
Sepp Hochreiter and Jurgen Schmidhuber.
\newblock 1997.
\newblock Long short-term memory.
\newblock {\it Neural computation}, 9(8):1735–1780.

\bibitem[\protect\citename{Mao \bgroup et al. \egroup}2014]{Mao:14}
Junhua Mao, Wei Xu, Yi Yang, Jiang Wang, and Alan L. Yuille.
\newblock 2014.
\newblock Explain images with multimodal recurrent neural networks.
\newblock {\it arXiv preprint arXiv:1410.1090}.

\bibitem[\protect\citename{Rolls and Deco}2002]{Rolls:02}
Edmund T. Rolls and Gustavo Deco.
\newblock 2002.
\newblock Computational Neuroscience of Vision.
\newblock Oxford University Press, USA.

\bibitem[\protect\citename{Hodosh \bgroup et al. \egroup}2013]{Hodosh:13}
Micah Hodosh, Peter Young, and Julia Hockenmaier.
\newblock 2013.
\newblock Framing image description as a ranking task: data, models and evaluation metrics.
\newblock {\it Journal of Artificial Intelligence Research}, 47: 853-899.

\bibitem[\protect\citename{Papineni \bgroup et al. \egroup}2002]{Papineni:02}
Kishore Papineni, Salim Roukos, Todd Ward, and Wei-Jing Zhu
\newblock 2002.
\newblock Bleu: a method for automatic evaluation of machine translation.
\newblock {\it Proceedings of the 40th annual meeting on association for computational linguistics}.

\bibitem[\protect\citename{Krizhevsky \bgroup et al. \egroup}2012]{Krizhevsky:12}
Alex Krizhevsky, Ilya Sutskever, and Geoffrey Hinton.
\newblock 2012.
\newblock ImageNet classification with deep convolutional neural networks.
\newblock In {\it NIPS}.

\bibitem[\protect\citename{Girshick \bgroup et al. \egroup}2014]{Girshick:14}
Ross Girshick, Jeff Donahue, Trevor Darrell, and Jitendra Malik
\newblock 2014.
\newblock Rich feature hierarchies for accurate object detection and semantic segmentation.
\newblock In {\it CVPR}.

\bibitem[\protect\citename{Mikolov \bgroup et al. \egroup}2013]{Mikolov:13}
Tomas Mikolov, Ilya Sutskever, Kai Chen, Greg Corrado, and Jeffrey Dean
\newblock 2013.
\newblock Distributed representations of words and phrases
and their compositionality.
\newblock In {\it NIPS}.

\bibitem[\protect\citename{Pennington \bgroup et al. \egroup}2014]{Pennington:14}
Jeffrey Pennington, Richard Socher, Christopher D. Manning
\newblock 2014.
\newblock Glove: Global vectors for word representation.
\newblock In {\it EMNLP}.

\bibitem[\protect\citename{Pascanu \bgroup et al. \egroup}2014]{Pascanu:14}
Razvan Pascanu, Caglar Gulcehre, Kyunghyun Cho, and Yoshua Bengio
\newblock 2014.
\newblock How to construct deep recurrent neural networks.
\newblock In {\it ICLR}.

\bibitem[\protect\citename{Hinton \bgroup et al. \egroup}2012]{Hinton:12}
Geoffrey Hinton, Nitish Srivastava, and Kevin Swersky.
\newblock 2012.
\newblock How to construct deep recurrent neural networks.
\newblock In {\it Lecture 6.5-rmsprop}.

\bibitem[\protect\citename{Zaremba \bgroup et al. \egroup}2014]{Zaremba:14}
Wojciech Zaremba, Ilya Sutskever, Oriol Vinyals
\newblock 2014.
\newblock Recurrent neural network regularization.
\newblock {\it arXiv preprint arXiv:1409.2329}.

\bibitem[\protect\citename{Szegedy \bgroup et al. \egroup}2014]{Szegedy:14}
Christian Szegedy, Wei Liu, Yangqing Jia, Pierre Sermanet, Scott Reed, Dragomir Anguelov, Dumitru Erhan, Vincent Vanhoucke, Andrew Rabinovich
\newblock 2014.
\newblock Going deeper with convolutions.
\newblock {\it arXiv preprint arXiv:1409.4842}.


\end{thebibliography}

\end{document}